\documentclass[a4paper]{article}
\usepackage{algorithm}
\usepackage{algpseudocode,spconf} 
\usepackage{graphicx,color}
\graphicspath{ {images/} }
\usepackage{amsmath, flushend}
\newtheorem{definition}{Definition}
\usepackage{multirow}
\usepackage{colortbl,hhline}

\usepackage[table]{xcolor}
\usepackage{pbox}
\usepackage{hyperref}

\newcommand{\win}{{\cellcolor[gray]{0.85}}}  
\newcommand{\loose}{{\cellcolor{red!15}}}  
\newcommand{\wwin}{{\cellcolor{green!15}}}  

\newcommand{\en}[2] {$#1 \pm #2$}
\newcolumntype{L}[1]{>{\raggedright\let\newline\\\arraybackslash\hspace{0pt}}m{#1}}
\newcolumntype{C}[1]{>{\centering\let\newline\\\arraybackslash\hspace{0pt}}m{#1}}
\newcolumntype{R}[1]{>{\raggedleft\let\newline\\\arraybackslash\hspace{0pt}}m{#1}}

\usepackage{fancyheadings}
\title{Learning Opposites Using Neural Networks}

\name{Shivam Kalra$^\dagger$, Aditya Sriram$^\dagger$,Shahryar Rahnamayan$^\ddagger$, H.R. Tizhoosh$^\dagger$}
\address{$^\dagger$ KIMIA Lab, University of Waterloo, Canada\\
   $^\ddagger$ Elect., Comp. \& Software Eng., University of Ontario Institute of Technology, Canada}
   
\begin{document}
\maketitle

\begin{abstract} Many research works have successfully extended algorithms such
  as evolutionary algorithms, reinforcement agents and neural networks using
  ``opposition-based learning'' (OBL). Two types of the ``opposites'' have been
  defined in the literature, namely \textit{type-I} and \textit{type-II}. The
  former are linear in nature and applicable to the variable space, hence easy to
  calculate. On the other hand, type-II opposites capture the ``oppositeness'' in
  the output space. In fact, type-I opposites are considered a special case of
  type-II opposites where inputs and outputs have a linear relationship. However,
  in many real-world problems, inputs and outputs do in fact exhibit a nonlinear
  relationship. Therefore, type-II opposites are expected to be better in
  capturing the sense of ``opposition'' in terms of the input-output relation. In
  the absence of any knowledge about the problem at hand, there seems to be no
  intuitive way to calculate the type-II opposites. In this paper, we introduce an
  approach to learn type-II opposites from the given inputs and their outputs
  using the artificial neural networks (ANNs). We first perform \emph{opposition
    mining} on the sample data, and then use the mined data to learn the
  relationship between input $x$ and its opposite $\breve{x}$. We have validated
  our algorithm using various benchmark functions to compare it against an
  evolving fuzzy inference approach that has been recently introduced. The results
  show the better performance of a neural approach to learn the opposites. This
  will create new possibilities for integrating oppositional schemes within
  existing algorithms promising a potential increase in convergence speed and/or
  accuracy.
\end{abstract}

\section{Introduction}
\label{sec:introduction} A large number of problems in engineering and science
are unapproachable with conventional schemes instead they are handled with
intelligent stochastic techniques such as evolutionary, neural network,
reinforcement and swarm-based algorithms. However, essential parameter for the
end users of aforementioned intelligent algorithms is to yield the solutions
within desirable accuracy in timely manner -- which remains volatile and
uncertain. Many heuristic methods exist to speed up the convergence rate of
stochastic algorithms to enhance their viability for complex real-world
problems. \textit{Opposite Based Computing (OBC)} is one such heuristic method
introduced by Tizhoosh in \cite{tizhoosh2005opposition}. The underlying idea is
simultaneous consideration of guess and opposite guess, estimate and opposite
estimate, parameter and opposite parameter \& so on in order to make more
educated decisions within the stochastic processes, that eventually results in
yielding solutions quickly and accurately.

In essence, learning the relationship between an entity and its opposite entity
for a given problem is a special case of \textit{a-priori knowledge}, which can
be beneficial for computationally intelligent algorithms in stochastic setups.
In context of machine learning algorithm, one may ask, why should effort be
spent on extraction of the opposite relations when input-output relationship
itself is not well defined? However, various research on this topic has shown
that simultaneous analysis of entities and their opposites can accelerate the
task in focus -- since it allows the algorithm to harness the knowledge about
symmetry in the solution domain thus allowing a better exploration of the
solutions. \textit{Opposition-based Differential Evolution (ODE)}, however,
seems to be the most successful oppositional inspired algorithm so far
\cite{rahnamayan2008opposition}.

There have been two types of opposites defined in literature \textbf{1)}
\textit{type-I} and \textbf{2)} \textit{type-II}. Generally, most learning
algorithms have an objective function; mapping the relationship between inputs
and their outputs -- which may be known or unknown. In such scenario, type-I
based learning algorithms deal with the relationship among input parameters,
based on their values, without considering their relationship with the objective
landscape. On contrary, type-II opposite requires a prior knowledge of the
objective function. Until 2015, all papers published on using OBL employed the
simple notion of type-I opposites which are conveniently, but naively defined on
the input space only, making a latent linearity assumption about the problem
domain. Tizhoosh and Rahnamayan \cite{tizhoosh2015learning} introduced the idea
of ``opposition mining'' and evolving rules to capture oppositeness in dynamic
environments. 

The paper is organized as follows: Section~\ref{sec:lit_rev} provides a
literature review on OBL. Section~\ref{sec:the_idea} introduces the idea to use
artificial neural network (ANN) to learn the opposites, and provides an overview
of type-I and type-II OBL. Finally, Section~\ref{sec:results} provides
experimental results and analysis and also a comparison of the proposed ANN
approach with the evolving fuzzy inference systems, a method recently proposed
in \cite{tizhoosh2015learning}.

\section{Background Review}
\label{sec:lit_rev} Roughly 10 years ago, the idea of opposition-based learning
(OBL) was introduced as a generic framework to improve existing learning and
optimization algorithms \cite{tizhoosh2005opposition}. This approach has
received a modest but growing attention by the research community resulting in
improving diverse optimization and learning techniques published in several
hundred papers \cite{al2010opposition}. A few algorithms have been reported to
employ ``oppositeness'' in their processing, including reinforcement learning
\cite{tizhoosh2005reinforcement,mahootchi2007opposition,tizhoosh2006opposition,mahootchi2014oppositional},
evolutionary algorithms
\cite{rahnamayan2006opposition,salehinejad2014type,rahnamayan2007novel,rahnamayan2008image},
swarm-based methods \cite{zhang2009novel,jabeen2009opposition,kaucic2013multi},
and neural networks \cite{rashid2010improved,yaghini2011hiopga,xu2014review}.

The majority of learning algorithms are tailored toward approximating functions
by arbitrary setting weights, activation functions, and the number of neurons in
the hidden layer. The convergence would be significantly faster towards the
optimal solution if these random initializations are close to the result
\cite{xu2014review}. On the contrary, if the initial estimates are far from the
optimal solution -- in the opposite corner of the search space, for instance,
then convergence to the ideal solution will take considerably more time or can
be left intractable \cite{ventresca2008diversity}. Hence, there is a need to
look simultaneously for a candidate solution in both current and opposite
directions to increase the convergence speed -- a learning mechanism denoted as
``opposition-based learning'' \cite{tizhoosh2005opposition}. The concept of OBL
has touched upon the various existing algorithm, and it has proven to yield
better results compared to the conventional method of determining the optimal
solution. A detailed survey on applications of OBL in soft-computing is
discussed by Al-Quanaieer et al. in \cite{al2010opposition}. The paper discusses
the integration of OBL when used for reinforcement learning, neural networks,
optimization, fuzzy set theory and fuzzy c-mean clustering. A review on each
algorithm states that applying OBL can be beneficial when applied in an
effective manner when applications use optimization algorithm, learning
algorithm, fuzzy sets and image processing.

Many problems in optimization involve minimization or maximization of some
scalar parameterized objective function, with respect to all its parameters. For
such problems, OBL can be used as a heuristic technique to quickly converge to
the solution within the search space by generating the candidates solutions that
have ``opposite-correlation'' instead of being entirely random.

The concept of ``opposite-correlation'' can be discussed from type-I and type-II
perspectives when an unknown function $y=f(x_1,x_2,\dots,x_n)$ needs to be
learned or optimized by relying on some sample data alone.

\begin{definition}
\label{def:typeI} Type-I opposite $\breve{x}_{I}$ of input $x$ is defined as
$\breve{x}_{I} = x_{\max} + x_{\min} - x$ where $x_{\max}$ is the maximum and
$x_{\min}$ is the minimum value of $x$.
\end{definition}

Computation of type-I opposites are easier due to its linear definition in the
variable space. On the contrary, type-II opposition scheme requires to operate
on the output space.

\begin{definition}
\label{def:typeII} Type-II opposite $\breve{x}_{II}$ of input $x$ is defined as
$\breve{x}_{II} = \left\{x_i | \breve{y}(x_i) = y_{min} + y_{max} - y(x_i)
\right\}$ where $y_{max} $ is the maximum value, and $y_{min}$ is the minimum
value of $y$.
\end{definition}

Type-II opposites may be difficult to incorporate in real-world problems because
\textbf{1)} their calculation may require a-priori domain knowledge, and
\textbf{2)} the inverse of the function $y$, namely $\breve{y}$, is not
available when we are dealing with unknown functions $y=f(x_1,x_2,\dots,x_n)$.
The focus of this paper is to develop a general framework to allow ANNs to learn
the relationship among the inputs and their corresponding type-II opposites.
Validation of the proposed approach is comprised of several benchmark functions.
Validation results on the benchmark functions demonstrate the effectiveness of
the proposed algorithm as an initial step for future developments in type-II OBL
approximations using neural networks.

\section{The Idea}
\label{sec:the_idea} Type-II (or true) opposite of $x$, denoted with
$\breve{x}_{II}$, is more intuitive when compared to type-I opposite in context
of ``non-linear'' functions. When looking at function $y = f(x_1, x_2, . . . ,
x_n)$ in a typical machine-learning setup, one may receive the output values $y$
for some input variables $x_1, x_2, \dots, x_n$. However, the function $y =
f(\cdot)$ itself is usually unknown otherwise there would be little
justification for resorting to machine-learning tools. Instead, one has some
sort of evaluation function $g(\cdot)$ (error, reward, fitness, etc.) that
enables to assess the quality of any guess $\hat{x_1}, \hat{x_2}, \dots ,
\hat{x_n}$ delivering an estimate $\hat{y}$ of the true/desired output $y$.

\begin{figure}[tb]
  \centering 
  \vspace{0.05in}
  \includegraphics[width=0.9\columnwidth]{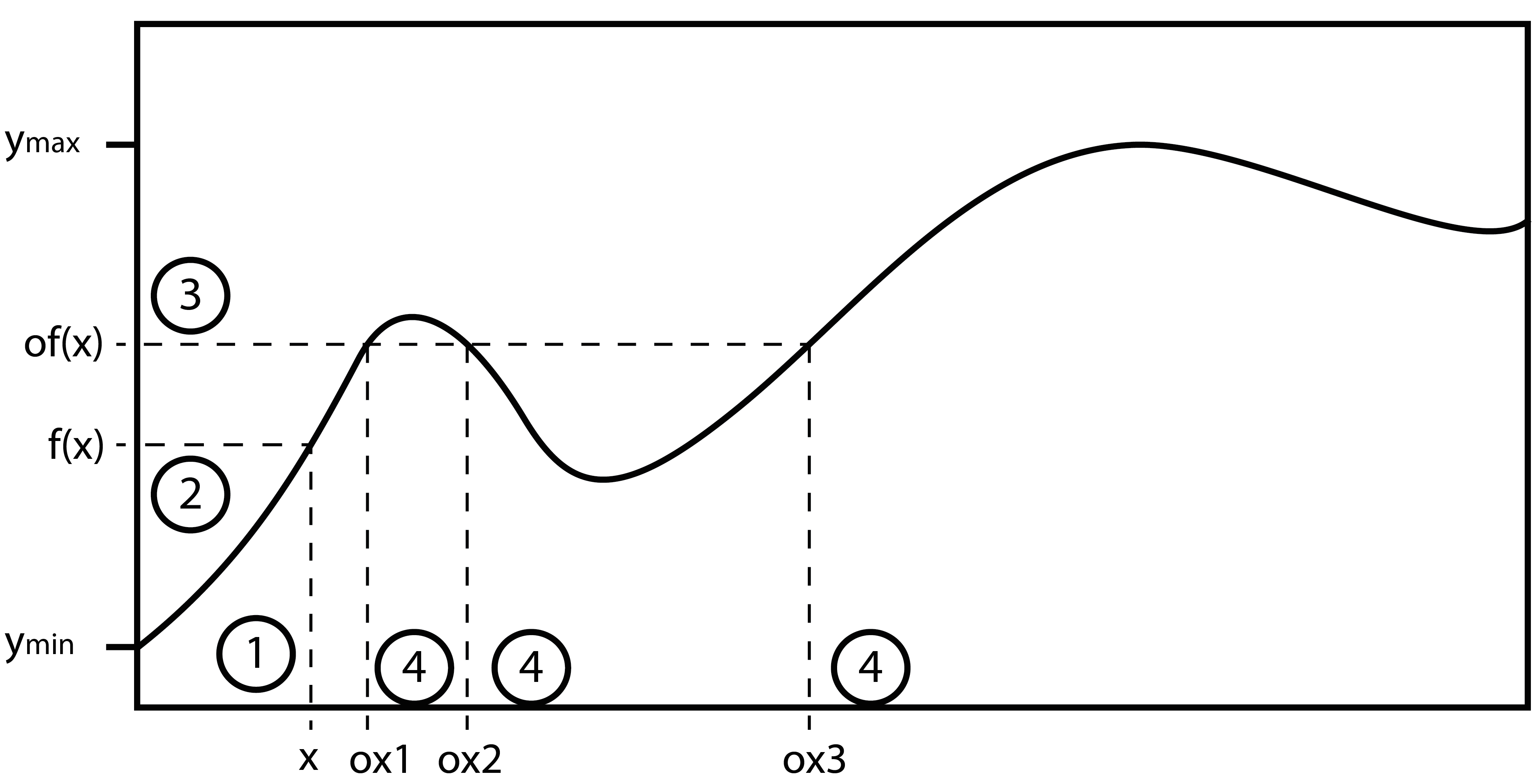} \caption{Type-II
    opposites [Adopted from \cite{tizhoosh2015learning}].} \label{fig:oblt2}
\end{figure}

Tizhoosh and Rahnamayan introduced the idea of \emph{opposition mining} as an
approach to approximate type-II opposites for training the learning algorithms
using fuzzy inference systems (FIS) with evolving rules in
\cite{tizhoosh2015learning}. Evolving FIS has received much attention lately
\cite{angelov2008evolving,angelov2008streams} which is now being used for
modeling nonlinear dynamic systems \cite{de2007line} and image classification
and segmentation \cite{lughofer2010line,othman2011evolving,othman2014efis}.
However, learning opposites with evolving rules are observed to be sensitive to
the parameters used and encounters difficulty in generalizing a large variety
of data. The proposed method in this paper uses opposition mining for training
artificial neural network to approximate the relationship between the input $x$
and its type-II opposite $\breve{x}_{II}$. This methodology can, of course, be
extended for various applications; hence, if a large training dataset is
available, then one can apply them at once instead of incremental changes. A
graphical representation of the type-II opposites, described in Definition
\ref{def:typeII}, is shown in Fig.~\ref{fig:oblt2}: \textbf{1)} Given variable
$x$, \textbf{2)} The corresponding $f(x)$ is calculated, \textbf{3)} from which
the opposite of $f(x)$, namely $of(x)$ is determined, and \textbf{4)} the
opposite(s) of $x$ are found to be: $ox1,ox2$ and $ox3$. However, there are
various challenges using this approach, which include: \textbf{1)} the output
range of $y = f(\cdot)$ may not be a-priori known, thereby an updated knowledge
on the output range of $[y_{min}, y_{max}]$ is needed, \textbf{2)} the precise
output of the type-II opposite may not be present in the given data; thereby, a
method to determine gaps in the dataset is needed to estimate the relationship
between the input and the output, and \textbf{3)} it is difficult to generalize
over a high dimension/range of data. In this paper, we put forward a concise
algorithm to learn opposites via neural networks -- which, as outlined in
Section~\ref{sec:results}, is proven to yield better results when compared to
recently introduced FIS approach.

\section{Learning Opposites} Neural networks with sufficient number of hidden
layers can learn any bounded degree polynomials with good accuracy
\cite{andoni2014learning}. Therefore, ANNs make a good candidate for learning
the nonlinear relationship between the inputs $x$ and their type-II opposites
$\breve{x}_{II}$.

In order to learn type-II opposites, we first need to sample the
(quasi-)opposites from the given input-output data. The first stage of our
algorithm is \emph{opposition mining}, which provides the data that can be
subsequently used by ANN for performing a nonlinear regression. One generally
assumes that the more data is available the better the approximation becomes for
$\breve{x}_{II} = f(x)$ as we have more (quasi-)opposites for training the ANN.
We assume that the range of input variable is known, $x_i \in
[{x^i}_{min},{x^i}_{max}]$ but the range of output, $y_j \in
[{y^j}_{min},{y^j}_{max}]$, may be apriori unknown. Since we are approximating
type-II opposites, we need to generate the (quasi-)opposite data from given
training data. Our approach consists of two distinct stages:\\
\textbf{Opposition Mining} -- The training data is sampled to establish the
output boundaries. Depending on a specific oppositeness scheme, all data points
in training data are processed to find (quasi-)opposites of each input. At the
end of the opposite mining, we have a corresponding (quasi-)opposite
(approximate of type-II opposite) for every input point in training data
as outlined in Algorithm \ref{alg:opmine}. There are different schemes for calculating the
opposition. Given a sample of random variable $x \in [x_{min}, x_{max}]$
with mean $\bar{x}$, the opposite of $x$ can be calculated as follows:
\begin{align} T_1: \quad \breve{x}_I &= z_{max} + x_{min} - x \\ T_2: \quad
  \breve{x}_I &= \left ( x + \frac{x_{min} + x_{max}}{2} \right ) \%\quad
                x_{max}\\ T_3: \quad \breve{x}_I &= 2\bar{x} - x
\end{align} In scheme $T_3$, calculated opposite $\breve{x}_{I}$ may go out of
the boundaries of the variable range. Therefore, for the experiments purposes,
we solved the boundary violation problem by switching the scheme to $T_1$
whenever necessary (Algorithm \ref{algo:oppmining}, Line
\ref{line:oppmineoverflow}). It is important to note that, opposites calculated
with any of the above schemes in output space when projected back on to the
variable space are known as type-II (true) opposites.

\begin{algorithm}[tb] \small
  \caption{Opposition Mining}\label{algo:oppmining} \textbf{Input:} $x$ inputs,
$y$ outputs \& $T_i$ opposition scheme \\ \textbf{Output:} type-II opposites
$\breve{x}_{II}$ in same order as input $x$
  \begin{algorithmic}[1] \Procedure{Opposition Mining}{$x, y, T_i$}

    \State{$y_{max} \gets max(y)$} \State{$y_{min} \gets min(y)$}
\State{$\bar{y} \gets mean(y)$}

    \For{$i \in [1, length(y)]$}
    
    \State{$y_I \gets y(i)$}

    \Comment{\textit{Choosing opposition scheme based on value of $T_i$}}
\If{$T_i = 0$} \State{$oppY \gets y_{max} + y_{min} - y_i$} \ElsIf{$T_I = 1$}
\State{$oppY \gets \left( y_I + \frac{y_{max} + y_{min}}{2} \right) \% y_{max}$}
\Else \State{$oppY \gets 2\bar{y} - y_i$}
    \EndIf{}

    \Comment{\textit{In case $oppY$ goes out of boundary range}} \If{$oppY
\notin \left[y_{max}, y_{min}\right]$} \label{line:oppmineoverflow} \State{$oppY
\gets y_{max} + y_{min} - y_i$}
    \EndIf{}

    \Comment{\textit{Getting index of element in $y$ closest to $oppY$}}
\State{$oppY_{idx} \gets argmin(|y - oppY|)$} \State{$\breve{x}_{II}(i) \gets
x(oppY_{idx})$}
    \EndFor{}
    
    \Return{$\breve{x}_{II}$}
    
    \EndProcedure{}
    
  \end{algorithmic}
  \label{alg:opmine}
\end{algorithm}

\textbf{Learning the Opposites} -- ANN is employed to approximate the
function $\breve{x}_{II} = f(x)$ that maps input and its type-II opposite. The
network is trained on the data collected from the opposition mining step, and
can be retrained progressively as more data comes in or it can be used to
predict the type-II opposites for a given input $x$. In the following sections,
we report the results of some experiments to verify the accuracy of our
algorithm, its superiority over existing FIS based technique and some discussion
on the usefulness of type-II opposites for machine-learning algorithms.

\section{Experiments and Results}
\label{sec:results} We have performed two experiment series to test the various
aspects of our algorithm including -- opposition mining, learning type-II using
ANN and prediction accuracy of the trained ANN model versus evolving fuzzy rules
model. The experiments deal with the approximation accuracy of type-II opposites
and application of type-II opposites for some standard optimization scenarios
respectively. For comparing the approximation accuracy, we used the 8 benchmark
functions for generating the data required for the opposition-mining and
subsequent training of ANN and evolving fuzzy rules models; which are taken from
\cite{tizhoosh2015learning}. It is important to note that, benchmarks function
have been intentionally kept simple and mostly monotonic in defined ranges, in
order avoid the surjective relationship of inputs and their (quasi-)opposites
during the opposition mining stage, thus allowing to extract the most feasible
patterns in the data to be used with learning algorithms. However, in order to
learn the type-II opposites across any general non-monotonic functions, it would
be required to decompose the function in question into monotonic piece-wise
ranges and subsequently perform the type-II approximation procedure on each of
the piece separately. We calculated the approximation error of our algorithm for
every benchmark function against different oppositeness schemes. We compared the
results against the recently published ones by approximating type-II using
evolving fuzzy rules \cite{tizhoosh2015learning}.

\begin{table*}[ht]
\tiny
  \centering
  \caption{Comparison between error in approximation of type-II opposite using
    trained ANN mode versus evolving fuzzy rules.}
  \label{tab:results}
  \begin{tabular}{|c|C{1cm}|C{2.7cm}|C{2.7cm}|C{1.3cm}|C{2.5cm}|} \hline
    \textbf{Benchmark Function} & \textbf{Opposition Scheme} & \textbf{Proposed Approach ($\mu \pm \sigma$)}
    & \textbf{Fuzzy based Approach ($\mu \pm \sigma$)} & \textbf{p-value} & \textbf{Welch's Test Results} \\ \hline

    \multirow{3}{*}{$f(x) = (2x + 8)^3$}
    & \win $T_1$ & \win \en{0.76}{0.85}  & \en{4.41}{2.44}   & 0.0005 & \wwin Proposed approach
    \\
    & $T_2$      & \en{9.53}{12.99}      & \en{11.96}{12.82} & 0.3398 & Not significant \\
    & $T_3$      & \en{4.65}{9.95}       & \en{6.82}{10.62}  & 0.3214 & Not significant \\ \hhline{======}

    \multirow{3}{*}{$f(x) = log(x+3)$}
    & $T_1$      & \win \en{18.95}{18.00}    & \en{30.05}{20.87} & 0.11 & \wwin Proposed approach $\approx$ 80\% confidence \\

    & $T_2$      & \en{10.02}{15.73}    & \en{11.20}{18.48} & 0.4398 & Not significant \\
    & \win $T_3$ & \en{2.98}{9.85} & \en{6.39}{7.87}   & 0.2020 & Not significant \\ \hhline{======}
  
    \multirow{3}{*}{$f(x) = 2*x$}
    & \win $T_1$ & \en{0.19}{0.26}   & \win \en{0.01}{0.01} & 0.9693 & \loose Fuzzy approach \\
    & $T_2$      & \en{6.24}{11.08}  & \en{21.03}{14.31} & 0.0091 & \wwin Proposed approach \\
    & $T_3$      & \en{0.30}{0.58}   & \en{0.25}{0.65} & 5759 & Not significant \\ \hhline{======}
 
   \multirow{3}{*}{$f(x) = x^2$}
    & \win $T_1$ & \win \en{0.49}{0.55}  & \en{3.04}{1.72} & 0.0005 & \wwin Proposed approach \\
    & $T_2$      & \en{8.46}{12.56}      & \en{15.02}{14.31} & 0.1456 & Not significant \\
    & $T_3$      & \en{3.61}{7.98}       & \en{4.41}{6.38} & 0.4039 & Not significant \\ \hhline{======}

    \multirow{3}{*}{$f(x) = \sqrt{x}$}
    & \win $T_1$ & \en{0.60}{1.15}      & \win \en{0.04}{0.13} & 0.9188 & \loose Fuzzy approach \\
    & $T_2$      & \en{2.91}{7.68}      & \en{18.99}{16.33} & 0.0074 & \wwin Proposed approach  \\
    & $T_3$      & \en{2.70}{5.91}      & \en{3.74}{4.22} & 0.3287 & Not significant \\ \hhline{======}

    \multirow{3}{*}{$f(x) = x^{3/2}$}
    & \win $T_1$ & \win \en{0.37}{0.31} & \en{1.68}{1.03} & 0.0014 & \wwin Proposed approach \\
    & $T_2$      & \en{5.16}{10.52}     & \en{17.89}{14.92} & 0.0212 & \wwin Proposed approach \\
    & $T_3$      & \en{2.15}{4.27}      & \en{2.66}{4.12} & 0.4022 & Not significant \\ \hhline{======}

    \multirow{3}{*}{$f(x) = x^3 + x^2 + 1$}
    & \win $T_1$ & \win \en{1.36}{2.84}   & \en{4.42}{2.72} & 0.0122 & \wwin Proposed approach \\
    & $T_2$      & \en{9.84}{12.57}       & \en{11.82}{12.95} & 0.3656 & Not significant \\
    & $T_3$      & \en{5.13}{10.79}       & \en{6.31}{9.99} & 0.4022 & Not significant \\ \hhline{======}
 
   \multirow{3}{*}{$f(x) = \frac{\sqrt{x+1}}{3}$}
    & \win $T_1$ & \en{1.63}{3.50}  & \win \en{0.06}{0.11} & 0.9061 & \loose Fuzzy approach \\
    & $T_2$      & \en{4.27}{8.17}  & \en{18.20}{16.79} & 0.0173 & \wwin Proposed approach \\
    & $T_3$      & \en{2.11}{5.15}  & \en{3.74}{4.61} & 0.2332 & Not significant \\ \hline

  \end{tabular}
  \label{tab:multicol}
\end{table*}

\subsection{Comparing with Evolving Fuzzy Rules} The results for 8 benchmark
functions (used in \cite{tizhoosh2015learning}) are summarized in
Table~\ref{tab:results}. Green cells represent the performance of ANN is
statistically significant with 95\% confidence (unless otherwise stated), whereas
red cells show the significance of evolving fuzzy rules based approach for the
respective opposition scheme. Cells marked gray represent the best results
achieved using any deployed method or opposition scheme for a given benchmark
function. The error in approximation of the type-II opposite
${\breve{x}_{II}}^*$ is inferred by comparing the value of the function at
approximated opposite $x_{II}$ and true opposite value of the function
${\breve{y}_{II}}^*$ at given input $x$. It is important to note that
${\breve{y}_{II}}^*$ can be calculated if input $x$, opposition scheme $T_i$ and
function $f$ are known:
\begin{equation}
  error(\breve{x}_{II}) \propto error(\breve{y}_{II}) = |\breve{y_{II}}^* - f(\breve{x}_{II})|
\end{equation}
The results are reported in Table~\ref{tab:multicol}. Overall $T_1$ seems to be
the best oppositeness scheme. As long as $y_{max}, y_{min}$ does not change for
the given sample data, $T_1$ and $T_2$ schemes allow continuous training of ANN.
The approximation seems to perform better overall except for the linear
functions and functions with square root power. The ANN approach seems to
generalize much better for the logarithm function at higher values of $x$ where
output changes much slower than change in $x$.

\begin{table*}[tb]
\tiny
  \caption{Errors values for Ackely, Booth and Bulkin test functions.}
  \label{tab:realworldprob}
  \centering
  \begin{tabular}{|c|c|c|c|c|}
    \hline
     Runs & $x_1^r$, $x_2^r$ & $x_{1,II,ANN}^r$, $x_{2,II,ANN}^r$ & $x_{1, II, FIS}^r$, $x_{2, II, FIS}^r$  & $x_{1, I}^r$, $x_{2, I}^r$\\ \hline 
    \multicolumn{5}{|c|}{\textbf{Ackley Function}} \\ \hline
    1. run & 487.70 $\pm$ 516.362    & 108.00 $\pm$ 139.03 & 117.43 $\pm$ 150.53 & 198.91 $\pm$ 222.677  \\
    2. run & 377.43  $\pm$ 455.85    & 108.65 $\pm$ 136.91 & 116.62 $\pm$ 138.02 & 155.24 $\pm$ 187.24 \\
    3. run & 512.31 $\pm$ 529.517    & 156.66 $\pm$ 162.61 & 143.05 $\pm$ 153.73 & 238.00 $\pm$ 246.19 \\
    4. run &  374.87 $\pm$ 388.63    & 124.20 $\pm$ 124.50 & 123.83 $\pm$ 134.19 & 154.55 $\pm$ 139.38 \\
    5. run &  415.84 $\pm$ 465.70    & 146.94 $\pm$ 157.14 & 128.90 $\pm$ 145.57 & 175.28 $\pm$ 218.05 \\ \hline

    \multicolumn{5}{|c|}{\textbf{Booth Function}} \\ \hline
 
    1. run & 424.53 $\pm$ 502.66     & 302.66 $\pm$ 274.64  & 337.39 $\pm$ 304.37  & 183.97 $\pm$ 197.39  \\   
    2. run & 420.03  $\pm$ 443.28    & 303.23 $\pm$ 318.19 & 330.74 $\pm$ 328.68 & 191.13 $\pm$ 166.01 \\
    3. run & 391.28 $\pm$ 445.53     & 318.19 $\pm$ 276.40 & 328.68 $\pm$ 280.56 & 166.01 $\pm$ 195.55 \\
    4. run & 338.94 $\pm$ 406.02    & 292.60 $\pm$ 290.77 & 123.83 $\pm$ 295.41 & 154.55 $\pm$ 174.13 \\
    5. run & 430.72 $\pm$ 496.36    & 323.67 $\pm$ 286.38 & 338.95 $\pm$ 289.79 & 187.19 $\pm$ 211.87 \\ \hline

    \multicolumn{5}{|c|}{\textbf{Bulkin Function}} \\ \hline
    
    1. run & 120.40 $\pm$  41.00    & 47.26 $\pm$ 18.99  & 63.89 $\pm$ 24.04  & 101.21 $\pm$ 38.46 \\
    2. run & 119.20  $\pm$ 50.42    & 53.78 $\pm$ 15.48 & 68.58 $\pm$ 26.75 & 97.11 $\pm$ 42.71 \\
    3. run & 126.78 $\pm$ 45.16    & 43.09 $\pm$ 17.24  & 72.83 $\pm$ 29.55 & 103.427 $\pm$ 38.58 \\
    4. run & 118.39 $\pm$ 47.31    & 49.17 $\pm$ 20.12 & 64.02 $\pm$ 30.43 & 94.27 $\pm$ 39.96 \\
    5. run & 128.58 $\pm$ 44.49    & 48.23 $\pm$ 20.96 & 66.43 $\pm$ 30.63 & 100.40 $\pm$ 38.16 \\ \hline

  \end{tabular}
\end{table*}

\subsection{Optimization Problems} 
In this experiment, we test three standard optimization functions which are
commonly used in literature of global optimization -- Ackely, Bulkin and Booth
functions \cite{tizhoosh2015learning}.

\textbf{Ackley Function} -- The Ackley function is given by
\begin{equation}
  \begin{split} f(x_1, x_2) &= 20 \left( 1 - \exp \left( -0.2 \sqrt{0.5(x_1^2 +
          x_2^2)} \right) \right) - \\ & \exp \big(0.5 * (cos(2 \pi x_1) + cos(2 \pi
    x_2))\big) + \exp(1)
  \end{split}\nonumber
\end{equation} The global minimum is $0$ at $(3, 0.5)$ with $x_1, x_2 \in
[-35, 35]$. 

\textbf{Bulkin Function} -- The Bulkin function is given by
\begin{equation}
  f(x_1, x_2) = 100 \sqrt{||x_2 - 0.01 x_1^2||} + 0.01||x_1 + 10|| \nonumber
\end{equation} 
The global minimum is $0$ at $(-10, 0)$ with $x_1 \in
[-15, -5]$ and $x_2 \in [-3, 3]$.

\textbf{Booth Function} -- The Booth function is given by
\begin{equation}
  f(x_1, x_2) = (x_1 +2x_2-7)^2 + (2x_1 + x_2 - 5)^2 \nonumber
\end{equation} The global minimum is $0$ at $(1, 3)$ such that $x_1, X_2 \in [-10, 10]$.

We train ANN and evolving fuzzy rules method for type-II opposites using $n_s =
1000$ samples from each of the benchmark function. This experiment enables to
verify three majors points: \textbf{1.)} test whether the fundamental statement
of OBL holds, namely that -- \textit{simultaneous consideration of a guess and
opposite guess provides faster convergence to the solution in learning and
optimization processes}, \textbf{2.)} test whether type-II opposites provides
any advantage over type-I, and \textbf{3.)} test if ANN based type-II
approximation provides any superiority over recently introduced evolving fuzzy
rule approach.

To conduct this experiment, we generate two random input samples $x_1^r$ and
$x_2^r$ and we calculate the error (distance from the global minimum). Then, we
approximate the opposites $\breve{x}_1^r$ and $\breve{x}_2^r$ and calculate the
error again. In order to find the solution, we chose the strategy that yields
less error and continues the process. We should expect to have a reduction or no
change in error at every iteration since we deliberately choose the outcome with
the least error. We carry out the experiment for both type-I $\breve{x}_I^r$ and
type-II opposites $\breve{x}_{II}^r$ generated by each of the candidate methods.
By recording the average error after $0.1*n_s$ iterations for multiple runs of
the experiments, we can test whether considering type-II opposite from either of
the candidate methods have any statistical significance over type-I in yielding
outcome closer to the global minimum. However, the focus is more on comparing
the two candidate approaches for approximating type-II (proposed and evolving
fuzzy rule-based), to see which one provides better estimates of type-II to
bring the optimization process closer to the global optima. The results for
Ackley, Booth, and Bulkin functions are shown in Table \ref{tab:realworldprob}.
The first column consists of random guesses, the second column contains type-II
opposite guesses estimated by ANN, and the third column contains type-II
opposite guesses using the FIS approach; the last column is type-I opposites. By
performing the t-test, results of type-II opposites obtained by ANN are
statistically significant compared to the FIS approach. Type-II is performing
better except for the Booth function.

\section{Conclusion} Ten years since the introduction of opposition-based
learning, the full potential of type-II opposites is still largely unknown. In
this paper, we put forward a method for learning type-II opposites with ANNs.
The core idea in this paper is to utilize the (quasi-)opposite data collected
from opposition-mining to learn the relationship between input $x$ and its
type-II opposite $\breve{x}_{II}$ using neural networks. We tested the proposed
algorithms with various benchmark functions and compared it against the existing
fuzzy rules-based approach. We showed the correctness of fundamental statement
of OBL scheme by utilizing type-II opposites on three of the famous global
optimization problems. One of the major hurdles for existing type-II
approximation methods (including proposed in this paper) is when the function in
question is highly non-monotonic or periodic in nature. In those circumstances,
the relationship between $x$ and $\breve{x}_{II}$ becomes surjective, causing
discontinuities in opposition mining. This makes it difficult for any learning
algorithm difficult to fit such discontinuous data. There is a potential for
improvement in future works where non-monotonic functions can be decomposed into
monotonic piece-wise intervals; each of the intervals can then be trained
separately.

\bibliographystyle{IEEEtran}
\bibliography{references}{}

\end{document}